# A Trident Quaternion Framework for Inertial-based Navigation Part I: Rigid Motion Representation and Computation

Wei Ouyang and Yuanxin Wu, *Senior Member*, *IEEE*

*Abstract*—Strapdown inertial navigation research involves the parameterization and computation of the attitude, velocity and position of a rigid body in a chosen reference frame. The community has long devoted to finding the most concise and efficient representation for the strapdown inertial navigation system (INS). The current work is motivated by simplifying the existing dual quaternion representation of the kinematic model. This paper proposes a compact and elegant representation of the body's attitude/velocity/position, with the aid of a devised trident quaternion tool in which the position is accounted for by adding a second imaginary part to the dual quaternion. Eventually, the kinematics of strapdown INS are cohesively unified in one concise differential equation, which bears the same form as the classical attitude quaternion equation. In addition, the computation of this trident quaternion-based kinematic equation is implemented with the recently proposed functional iterative integration approach. Numerical results verify the analysis and show that incorporating the new representation into the functional iterative integration scheme achieves high inertial navigation computation accuracy as well.

## I. INTRODUCTION

THE transition from gimbaled inertial navigation to strapdown inertial navigation has spurred a wealth of frameworks and algorithms over the last half century to achieve an accurate solution of attitude, velocity and position of a rigid body [1]-[7]. In a strapdown inertial navigation system (INS), a triad of gyro outputs are used to propagate the attitude with respect to a reference frame, which is further exploited to calculate the velocity/position with a triad of accelerometer outputs [8].

Appropriate parameterization of the body attitude is vital to the INS mechanism and the subsequent navigation computation process. A number of attitude representations have been proposed for strapdown attitude computation [9]. Among them the Euler angle, the rotation vector, the direction cosine matrix (DCM), and the quaternion are routinely used in the navigation and control of land/underwater vehicles, airplanes, spacecraft and robots, etc [10], [11]. The fundamental object of attitude computation is to solve the ordinary differential equations in those parameters [12], [13]. Specifically, the three-dimension rotation vector-based attitude parameterization is believed to be minimal and free from the intrinsic constraints of the quaternion and DCM. Despite the unit-norm constraint, the quaternion parameterization has been widely accepted in the navigation community owing to its benefits of non-singularity, succinctness and the linearity in its differential equation for efficient algebraic computation [14]-[17]. This work primarily considers the quaternion-related description of the strapdown INS.

Since initially developed by Hamilton in 1843 [18], the quaternion has emerged as a powerful and efficient tool in representing the attitude kinematics and dynamics in many fields. Further efforts have been devoted to enriching the quaternion algebra family by developing new algebraic structures. The description of spatial rotation and translation of a rigid body was initially studied in the screw theory developed by Ball in 1900 [19]. This combined motion was further described by the dual quaternion, which is believed as the most elegant and compact mathematical description of rigid body motions [20]. Indeed, the dual quaternion is by definition the combination of the dual number and the quaternion [21]. In 1992, Branets and Shmyglevsky [22] firstly proposed to use dual quaternions to describe the attitude and translation simultaneously in the strapdown INS, where three continuous kinematic equations in dual quaternions were formulated to express the kinematics of strapdown INS. To the author's knowledge, researches and applications of the dual quaternion are somewhat tepid in the navigation community until our group further investigated the dual quaternion formulation for the strapdown INS algorithm design in [23] and derived linearized error models in terms of dual quaternions in [24]. However, the inherent redundancies of three dual quaternion kinematic equations impeded their applications in strapdown INS computation and integrated navigation estimation.

In recent twenty years, especially after our group's work [23], booming studies have revived in applying dual quaternions to kinematics and dynamics of mechanics and robotics, manipulator control and state estimation [25]-[30]. For instance, Wang exploited the geometric structure of the dual quaternion in kinematic modeling and control law design for position

The paper was supported in part by National Key R&D Program of China (2018YFB1305103) and National Natural Science Foundation of China (61673263). A short version was submitted to International Conference on Integrated Navigation Systems, Saint Petersburg, Russia, 2021.

W. Ouyang and Y. Wu are with Shanghai Key Laboratory of Navigation and Location-based Services, School of Electronic Information and Electrical Engineering, Shanghai Jiao Tong University, Shanghai 200240, China (email: ywoulife@sjtu.edu.cn, yuanx_wu@hotmail.com).

tracking [26] and formation coordination [27]. Filipe applied the dual quaternion representation in spacecraft velocity-free pose tracking control [29] and spacecraft pose estimation based on the extended Kalman filter [30]. Similarly, the unit dual quaternion was used in [31]-[34] to design Kalman/particle filters for robotics rigid registration and motion estimation. More recently, in the realm of modeling multiple-chain/body kinematics and dynamics, two interesting dual algebra structures are invented, i.e., the hyper-dual and multi-dual algebra. Initially, the hyper-dual numbers and corresponding characteristics were developed by Fike [35], [36] to achieve exact second-derivative calculations, free from both truncation and subtractive cancellation errors. Subsequently, the hyper-dual numbers were applied to model multibody kinematics and dynamics succinctly [37], taking advantage of the "automatic differentiation" feature. Notably, the hyper-dual quaternion proposed by Cohen [38] incorporated both the body pose (attitude and position) and body velocity compactly without the need to take further pose differentiation. By contrast, the multi-dual numbers and their algebraic characteristics developed by Farid in [39] were recently used in studying higher-order acceleration field properties of general rigid body motions [40].

Inspired by the aforementioned seminal works [35], [39], we endeavor to find the most concise and efficient representation of the strapdown INS by minimizing the redundancy in previous works [23], [24]. A new algebraic structure, named the trident quaternion, is developed to simplify the kinematic equations of the strapdown INS. Specifically, the former three kinematic equations in dual quaternions [22], [23] are successfully united in one compact trident quaternion based kinematic equation, which also takes a similar form as the classical quaternion attitude equation.

As for the strapdown INS computation, Jordan [2] and Bortz [3] laid the foundation of modern-day strapdown INS algorithm framework, featured in using the approximated rotation vector to update the attitude incrementally. The velocity computation takes the first-order approximation of the body attitude in integrating the transformed specific force [6]. However, for future high-dynamic applications or ultra-high precision cold-atom gyroscopes, the current algorithms might cause accuracy compromise due to the above fundamental approximations. Quite recent works [45], [46] by Ignagni, one of the strapdown INS system pioneers [47], illuminates the inherent accuracy limitations of traditional algorithms and consequently proposes a couple of algorithmic enhancement. The functional iterative integration approach recently proposed in [12], [44] for the first time reduces the coning/sculling/scrolling non-commutativity errors to almost the machine precision, which hopefully would best benefit those applications where the computational errors take a significant percentage relative to sensor/gravity errors in the strapdown INS system, e.g., under sustained highly dynamic motions [45], [46]. In view of future applications, this work also tries to solve the obtained trident quaternion kinematic equation by the functional iterative integration approach.

The main contributions of this work reside in: 1) the redundancies in previous dual quaternion kinematic equations [22], [23] are analyzed and minimized; 2) a new trident quaternion algebra is developed to concisely describe the strapdown INS kinematics, which leads to one unified trident quaternion differential equation; 3) the functional iterative approach combined with Chebyshev polynomial approximation is applied to accurately solve the trident quaternion kinematic equation to the machine precision.

The remaining content is structured as follows: Section II provides the mathematical preliminary of kinematic representations of a rigid body. Section III devotes to simplifying the redundant dual quaternion kinematic equations. Subsequently, the trident numbers and trident quaternions are developed in Section IV to represent the kinematics of strapdown INS. Section V presents the process of solving the trident quaternion kinematics with the functional iterative integration. Section VI numerically verifies the proposed framework and the computational scheme. Finally, the conclusion is provided in Section VII.

## II. MATHEMATICAL PRELIMINARIES

This section provides a brief summary of the quaternion and dual quaternion. Their operations and properties in detail can be found in [23] and [29].

### A. Quaternion

Quaternions are generalized complex numbers invented by Hamilton to represent rotations in the three-dimensional space. A quaternion is defined as $q = [s, \mathbf{v}]$ or $q = s + ix + jy + kz$, where $\mathbf{v} = [x, y, z]^T \in \mathbb{R}^3$ is the vector part and $s, x, y, z$ are real scalars. $i, j, k$ are the generalized imaginary units, which satisfy $i^2 = j^2 = k^2 = -1$, $ij = -ji = k$, $ki = -ik = j$, and $jk = -kj = i$. Note that in this work only the three-dimensional vectors are denoted in bold font.

By definition, the addition and multiplication rules of two quaternions are given as

$$q_1 + q_2 = [s_1 + s_2, \mathbf{v}_1 + \mathbf{v}_2]$$
$$q_1 \circ q_2 = [s_1 s_2 - \mathbf{v}_1 \mathbf{v}_2, s_1 \mathbf{v}_2 + s_2 \mathbf{v}_1 + \mathbf{v}_1 \times \mathbf{v}_2] \quad (1)$$

where '$\times$' is the vector cross product, and '$\circ$' is the quaternion product.

The norm of a quaternion is computed as $\|q\| = q \circ q^*$. Therefore, the conjugate of a unit quaternion is given as $q^* = q^{-1} = [s, -\mathbf{v}]$.

Regarding the rotation of the frame $O$ to the frame $N$ about a unit vector $\mathbf{n}$ with an angle $\theta$, the unit quaternion $q_{ON} = [\cos(\theta/2), \sin(\theta/2)\mathbf{n}]$ transforms the vector coordinate in the original frame $O$ to that in another frame $N$ as

$$r^N = q_{ON}^* \circ r^O \circ q_{ON} = q_{NO} \circ r^O \circ q_{NO}^* \quad (2)$$

where $r^N, r^O$ are quaternions with zero scalar part, e.g., the vector quaternion $r^N = [0, \mathbf{r}^N]$ and $q_{ON}^* = q_{NO}$.

The quaternion kinematic equation is defined as

$$2\dot{q}_{ON} = q \circ \omega_{ON}^N = \omega_{ON}^O \circ q_{ON} \quad (3)$$

where $\omega_{ON}^N = [0, \boldsymbol{\omega}_{ON}^N]$ and the vector part $\boldsymbol{\omega}_{ON}^N$ is the angular rate vector of the frame $N$ w.r.t. the frame $O$, expressed in the frame $N$.

*B. Dual Quaternion*

The dual quaternion is by definition the combination of dual number theory with quaternions. Dual quaternions can be regarded as quaternions with dual parts, i.e., $\hat{q} = [\hat{s}, \hat{\mathbf{v}}]$, where $\hat{s} = s + \varepsilon s'$ is the dual number and $\hat{\mathbf{v}} = \mathbf{v} + \varepsilon \mathbf{v}'$ is a dual vector [23]. The dual unit $\varepsilon$ is defined by $\varepsilon^2 = 0$ and $\varepsilon \neq 0$.

Moreover, a dual quaternion can be explicitly reformulated as a dual number with quaternions as the real and dual part, respectively.

$$\hat{q} = q + \varepsilon q' \qquad (4)$$

where both $q$ and $q'$ are usual quaternions.

In this work, the dual quaternions with vanishing scalar part in each quaternion are named as dual vector quaternions, which are indiscriminate with usual dual quaternions.

The addition and multiplication between dual quaternions are defined as

$$\begin{aligned}\hat{q}_1 + \hat{q}_2 &= q_1 + q_2 + \varepsilon(q_1' + q_2') \\ \hat{q}_1 \circ \hat{q}_2 &= q_1 \circ q_2 + \varepsilon(q_1 \circ q_2' + q_1' \circ q_2)\end{aligned} \qquad (5)$$

Similarly, the conjugate of a dual quaternion is denoted as $\hat{q}^* = q^* + \varepsilon q'^*$ or $\hat{q}^* = [\hat{s}, -\hat{\mathbf{v}}]$. The norm of a dual quaternion is $\|\hat{q}\| = \hat{q} \circ \hat{q}^*$ and its inverse is equivalent to its conjugate for a unit dual quaternion.

Dual quaternion is usually used to represent the rotation and translation between two frames simultaneously. The relative motion of the frame $N$ w.r.t. the frame $O$ can be described by a rotation $q_{ON}$ followed by a translation $\mathbf{t}^N$ or a translation $\mathbf{t}^O$ followed by a rotation $q_{ON}$. The unit dual quaternion defined to describe this relationship is

$$\hat{q}_{ON} = q_{ON} + \varepsilon q'_{ON} = q_{ON} + \varepsilon \frac{1}{2} t^O \circ q_{ON} = q_{ON} + \varepsilon \frac{1}{2} q_{ON} \circ t^N \qquad (6)$$

in which $2q'_{ON} = t^O \circ q_{ON} = q_{ON} \circ t^N$ and $t^O = [0, \mathbf{t}^O]$, $t^N = [0, \mathbf{t}^N]$.

Similar to the kinematic equation of quaternions, the kinematic equation of dual quaternion is formulated as

$$2\dot{\hat{q}}_{ON} = \hat{q}_{ON} \circ \hat{\omega}_{ON}^N = \hat{\omega}_{ON}^O \circ \hat{q}_{ON} \qquad (7)$$

where the dual vector quaternions $\hat{\omega}_{ON}^N$ and $\hat{\omega}_{ON}^O$, also called twists, are further denoted as [23]

$$\begin{aligned}\hat{\omega}_{ON}^N &= \omega_{ON}^N + \varepsilon(\dot{t}^N + \omega_{ON}^N \times t^N) \\ \hat{\omega}_{ON}^O &= \omega_{ON}^O + \varepsilon(\dot{t}^O + t^O \times \omega_{ON}^O)\end{aligned} \qquad (8)$$

where vectors are automatically transformed to the corresponding vector quaternions in computations with quaternions.

## III. Dual Quaternion and Vector Hybrid Representation

The dual quaternion formulation of inertial navigation was presented in [22]-[24], where three differential equations of dual quaternion were formulated to represent the principle of inertial navigation. However, there are considerable redundancies in these equations as it uses twenty-four states to encode the rigid motion with nine degrees of freedom (three for each of attitude, velocity and position). Next we will show that the redundancies could be significantly reduced by a novel formulation of dual quaternion and vector hybrid model. Here, we start with similar thought of [23] and devote to simplifying the previous dual quaternion representation of strapdown INS kinematic equations.

Consider the inertial frame (*i*-frame), the WGS-84 Earth frame (*e*-frame) and the IMU body frame (*b*-frame). $\mathbf{r}^e$ denotes the position vector of the IMU relative to the earth. The position coordinates in *i*-frame and *e*-frame are related by

$$\mathbf{r}^i = \mathbf{C}_e^i \mathbf{r}^e \qquad (9)$$

Note that for normal navigation applications near the earth, the earth revolution is typically not considered [9], except for long-distance/long-time navigation scenarios. The superscripts denote the frames where to express the vector and $\mathbf{C}_e^i$ is the attitude matrix of *i*-frame with respect to *e*-frame.

The attitude kinematic equation is given by

$$\dot{\mathbf{C}}_e^i = \mathbf{C}_e^i (\boldsymbol{\omega}_{ie}^e \times) \qquad (10)$$

where $\boldsymbol{\omega}_{ie}^e = [0, 0, \omega_{ie}^e]^T$ is the constant earth rotation angular velocity in *e*-frame and $\boldsymbol{\omega}_{ie}^e \times$ is the antisymmetric matrix formed by this vector. Taking the time derivative of Eq. (9) and using Eq. (10),

$$\dot{\mathbf{r}}^i = \mathbf{C}_e^i (\boldsymbol{\omega}_{ie}^e \times) \mathbf{r}^e + \mathbf{C}_e^i \dot{\mathbf{r}}^e \qquad (11)$$

Take a further time derivative of Eq. (11) and define $\mathbf{v}^e \triangleq \dot{\mathbf{r}}^e$

$$\mathbf{f}^i + \mathbf{g}^i = \ddot{\mathbf{r}}^i = \mathbf{C}_e^i (\boldsymbol{\omega}_{ie}^e \times)^2 \mathbf{r}^e + 2\mathbf{C}_e^i (\boldsymbol{\omega}_{ie}^e \times) \mathbf{v}^e + \mathbf{C}_e^i \dot{\mathbf{v}}^e \qquad (12)$$

where $\mathbf{f}^i$ denotes the specific force by the applied non-gravitational forces and $\mathbf{g}^i$ is the gravitational acceleration.

From Eq. (12), the vector

$$\dot{\mathbf{r}}^i = \dot{\mathbf{r}}^i(0) + \int_0^t (\mathbf{f}^i + \mathbf{g}^i) d\tau \qquad (13)$$

where $\dot{\mathbf{r}}^i(0)$ is the initial velocity of the IMU in *i*-frame at time zero.

As the specific force is measured in *b*-frame onboard the vehicle and the gravitational acceleration is usually provided in *e*-frame or some local-level frames, Equation (12) can be rewritten as the traditional velocity equation in *e*-frame by multiplying $\mathbf{C}_i^e$ on both sides,

$$\dot{\mathbf{v}}^e = \mathbf{C}_b^e \mathbf{f}^b - 2\boldsymbol{\omega}_{ie}^e \times \mathbf{v}^e + \mathbf{g}_l^e \qquad (14)$$

where the term $\mathbf{g}_l^e = \mathbf{g}^e - (\boldsymbol{\omega}_{ie}^e \times)^2 \mathbf{r}^e$ is collectively known as the local gravity vector, and $\mathbf{C}_b^e$ is the attitude matrix of *e*-frame with respect to *b*-frame and satisfies the kinematics as

$$\dot{\mathbf{C}}_b^e = \mathbf{C}_b^e (\boldsymbol{\omega}_{eb}^b \times) = \mathbf{C}_b^e (\boldsymbol{\omega}_{ib}^b \times) - (\boldsymbol{\omega}_{ie}^e \times) \mathbf{C}_b^e \qquad (15)$$

where $\boldsymbol{\omega}_{ib}^b$ is the body angular velocity measured by gyroscopes. Equations (14)-(15), together with the trivial definition $\dot{\mathbf{r}}^e = \mathbf{v}^e$, constitute the routine navigation equations in *e*-frame [9]. In specific, they respectively characterize the translation and rotation of the vehicle. The attitude matrix in (15) could be alternatively expressed in attitude quaternion.

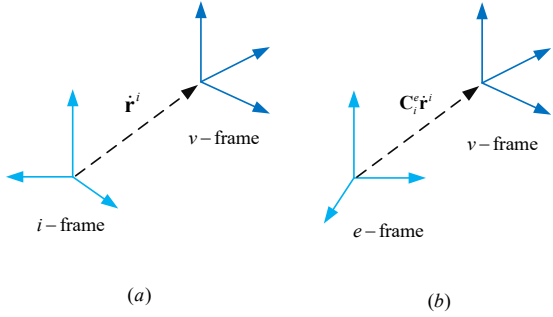

Figure 1: The relationship of the newly defined "total velocity" *v*-frame with *i*-frame and *e*-frame. (a) the *v*-frame is displaced from the *i*-frame with the vector $\dot{\mathbf{r}}^i$; (b) the *v*-frame is displaced from the *e*-frame with the vector $\mathbf{C}_i^e \dot{\mathbf{r}}^i$.

Notably, the dual quaternion provides a unified and efficient way to encode both rotational and translational motion between two frames. Its property is very much like that of quaternion for attitude. In order to simplify the dual quaternion representations in [22], [23], first define a new *total velocity* frame (*v*-frame) which aligns with *b*-frame in orientation but displaces relative to *i*-frame by the velocity vector $\dot{\mathbf{r}}^i$, as illustrated in Fig. 1(a). Since *e*-frame is related with *i*-frame with the rotation matrix $\mathbf{C}_i^e$, the translation between *e*-frame and *v*-frame becomes $\mathbf{C}_i^e \dot{\mathbf{r}}^i$, as shown in Fig. 1(b).

Intuitively, the abstract concept of "translation" can be perceived as the relative displacement between two frames, which is happened to be the velocity vectors here. Note that in Eqs. (9)-(15), $\dot{\mathbf{r}}^i$ is the displaced vector of the IMU relative to *i*-frame and $\mathbf{C}_i^e \dot{\mathbf{r}}^i$ expresses this vector in *e*-frame. It relates to the usual *e*-frame velocity by $\mathbf{C}_i^e \dot{\mathbf{r}}^i = \boldsymbol{\omega}_{ie}^e \times \mathbf{r}^e + \mathbf{v}^e$, according to Eq. (11). In this regard, the "translation" here is a local and abstract concept between two considered frames [23].

According to Eq. (6), the dual quaternion used to describe the relationship between *v*-frame and *e*-frame is given as

$$\hat{q}_{ev} \triangleq q_{ev} + \varepsilon q'_{ev} = q_{eb} + \varepsilon \frac{1}{2}\left(\mathbf{C}_i^e \dot{\mathbf{r}}^i\right) \circ q_{eb} \quad (16)$$

Note that the vector $\mathbf{C}_i^e \dot{\mathbf{r}}^i$ in Eq. (16) is automatically used as a vector quaternion when multiplying with another quaternion.

Taking the time derivative of Eq. (16), the kinematic equation of the dual quaternion is given by (see detailed derivations in Appendix A)

$$2\dot{\hat{q}}_{ev} = \hat{q}_{ev} \circ \hat{\omega}_{ib}^b - \hat{\omega}_{ie}^e \circ \hat{q}_{ev} \quad (17)$$

The above two dual vector quaternions (also called as twists) are given as

$$\begin{aligned} \hat{\omega}_{ib}^b &= \omega_{ib}^b + \varepsilon f^b \\ \hat{\omega}_{ie}^e &= \omega_{ie}^e - \varepsilon g^e \end{aligned} \quad (18)$$

where $\hat{\omega}_{ib}^b$ is exactly the twist for the total velocity frame, and $\hat{\omega}_{ie}^e$ is exactly the twist for the gravitational velocity frame in [23]. As a matter of fact, Eq. (17) can be alternatively obtained from combining the two previously derived dual quaternions equations in [23] yet by a minor adjustment of frame definitions (see Appendix B for details). It is a nice property that $\hat{\omega}_{ib}^b$ or $\hat{\omega}_{ie}^e$ can be readily built from direct measurements (the gyro outputs $\boldsymbol{\omega}_{ib}^b$ and the accelerometer outputs $\mathbf{f}^b$) or the earth model (the earth rotation angular velocity $\boldsymbol{\omega}_{ie}^e$ and the gravitational acceleration $\mathbf{g}^e$).

From Eq. (11), the IMU position relative to *e*-frame can be represented by the dual quaternion $\hat{q}_{ev}$ as

$$\dot{\mathbf{r}}^e = \mathbf{C}_i^e \dot{\mathbf{r}}^i - \boldsymbol{\omega}_{ie}^e \times \mathbf{r}^e = 2q'_{ev} \circ q^*_{ev} - \boldsymbol{\omega}_{ie}^e \times \mathbf{r}^e \quad (19)$$

To summarize, the inertial navigation equation has been so far represented by a dual quaternion and vector (DQv) hybrid model, as given in Eqs. (17) and (19). Specifically, the former encodes the attitude and the velocity by way of a dual quaternion, and the latter directly uses the common position vector. Compared against the original dual quaternion formulation in [22], [23], the state dimension has been reduced from twenty four to eleven now. The DQv's information flow is plotted in Fig. 2 and compared with that of the conventional *e*-frame formulation [9]. The attitude and velocity (six degrees of freedom) are now entangled together by the eight-dimensional dual quaternion. This section paves the way for more efficient and elegant representation of kinematics in Section IV.

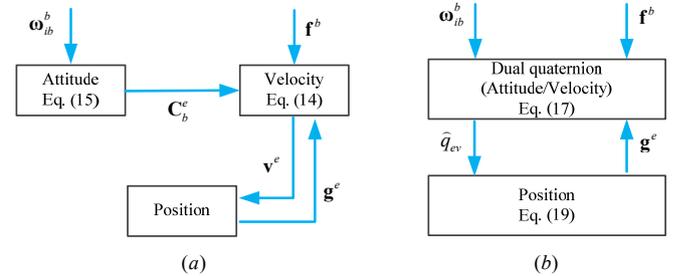

Figure 2: Information flow in (a) typical Earth-frame mechanization and in (b) the derived dual quaternion & vector model. Arrowed lines indicate information flow directions and the associated symbols mean that their computation needs to feed on the source information.

IV. TRIDENT QUATERNION REPRESENTATION

The former section has simplified the strapdown INS kinematic representation from three dual quaternion kinematic equations to one dual quaternion kinematic equation and one vector equation. However, it can be observed that the position vector equation cannot be absorbed into the dual quaternion equation. This is mainly caused by the fact that only one "translation" is defined in the dual quaternion theory, while the spatial motion of a rigid body involves relative velocity "translation" and relative position "translation" w.r.t. a reference frame. To address this problem, a new algebraic structure is developed here to achieve more succinct representation of the strapdown INS kinematics.

## A. Trident Quaternion and Its Properties

*Definition 1.* The trident dual numbers are defined as $\breve{a} = a_0 + \varepsilon_1 a_1 + \varepsilon_2 a_2$, where $a_0, a_1, a_2 \in \mathbb{R}$, $\varepsilon_1^2 = \varepsilon_2^2 = \varepsilon_1 \varepsilon_2 = 0$, $\varepsilon_1 \neq 0$ and $\varepsilon_2 \neq 0$.

Theoretically, the trident dual numbers are direct extension of the dual number by juxtaposing a second imaginary part. The two imaginary parts are equivalent in algebraic operations. Similarly, we can define the arithmetic operations of trident numbers like the dual numbers as follows

$$\begin{aligned} \breve{a} + \breve{b} &= a_0 + b_0 + \varepsilon_1 (a_1 + b_1) + \varepsilon_2 (a_2 + b_2) \\ \lambda \breve{a} &= \lambda a_0 + \lambda \varepsilon_1 a_1 + \lambda \varepsilon_2 a_2 \\ \breve{a}\breve{b} &= a_0 b_0 + \varepsilon_1 (a_0 b_1 + a_1 b_0) + \varepsilon_2 (a_0 b_2 + a_2 b_0) \end{aligned} \quad (20)$$

where $\lambda$ is a real scalar. Alternatively, trident vectors can be accordingly defined as trident numbers, except that their real part and imaginary parts are usual vectors.

A trident quaternion is devised as follows by combining quaternions with trident numbers.

*Definition 2.* The trident quaternion is defined as $\breve{q} = q + \varepsilon_1 q' + \varepsilon_2 q''$, where $q_0, q_1, q_2$ are usual quaternions and the imaginary units satisfy $\varepsilon_1^2 = \varepsilon_2^2 = \varepsilon_1 \varepsilon_2 = 0$, $\varepsilon_1 \neq 0$ and $\varepsilon_2 \neq 0$.

For any two trident quaternions $\breve{q}_1 = q_1 + \varepsilon_1 q_1' + \varepsilon_2 q_1''$ and $\breve{q}_2 = q_2 + \varepsilon_1 q_2' + \varepsilon_2 q_2''$, the arithmetic operations naturally follow as

$$\begin{aligned} \breve{q}_1 + \breve{q}_2 &= q_1 + q_2 + \varepsilon_1 (q_1' + q_2') + \varepsilon_2 (q_1'' + q_2'') \\ \lambda \breve{q} &= \lambda q + \lambda \varepsilon_1 q' + \lambda \varepsilon_2 q'' \\ \breve{q}_1 \circ \breve{q}_2 &= q_1 \circ q_2 + \varepsilon_1 (q_1 \circ q_2' + q_1' \circ q_2) + \varepsilon_2 (q_1 \circ q_2'' + q_1'' \circ q_2) \end{aligned} \quad (21)$$

As for the investigated inertial navigation system, in which the total velocity frame is defined relative to the *e*-frame, the trident quaternion is defined as

$$\breve{q}_{eb} = q_{eb} + \varepsilon_1 q_{eb}' + \varepsilon_2 q_{eb}'' \quad (22)$$

in which, we directly replace *v* with *b* in contrast with Eq. (16).

Similar to the derivations in Section III, the second imaginary part extended here is used to encompass the position translation of the total velocity frame w.r.t. the *e*-frame. Hence, the trident quaternion is finally given as

$$\breve{q}_{eb} = q_{eb} + \varepsilon_1 \frac{1}{2}(\mathbf{C}_i^e \dot{\mathbf{r}}^i) \circ q_{eb} + \varepsilon_2 \frac{1}{2}(\mathbf{C}_i^e \mathbf{r}^i) \circ q_{eb} \quad (23)$$

Taking the derivative of Eq. (23) and using (50), the kinematic equation is obtained as

$$2\dot{\breve{q}}_{eb} = q_{eb} \circ \omega_{eb}^b + \varepsilon_1 \begin{bmatrix} \frac{1}{2}(\mathbf{C}_i^e \dot{\mathbf{r}}^i) \circ q_{eb} \circ \omega_{ib}^b + q_{eb} \circ f^b \\ + g^e \circ q_{eb} - \frac{1}{2}\omega_{ie}^e \circ (\mathbf{C}_i^e \dot{\mathbf{r}}^i) \circ q_{eb} \end{bmatrix} \\ + \varepsilon_2 \begin{bmatrix} \frac{1}{2}(\mathbf{C}_i^e \mathbf{r}^i) \circ q_{eb} \circ \omega_{ib}^b - \frac{1}{2}\omega_{ie}^e \circ (\mathbf{C}_i^e \mathbf{r}^i) \circ q_{eb} \\ + (\mathbf{C}_i^e \dot{\mathbf{r}}^i) \circ q_{eb} \end{bmatrix} \quad (24)$$

which can be further formulated in a beautiful and compact form

$$2\dot{\breve{q}}_{eb} = \breve{q}_{eb} \circ \breve{\omega}_{ib}^b - \breve{\omega}_{ie}^e \circ \breve{q}_{eb} \quad (25)$$

Referring to the representations in Section III, two *trident twists* could be accordingly defined as

$$\begin{aligned} \breve{\omega}_{ib}^b &= \omega_{ib}^b + \varepsilon_1 f^b \\ \breve{\omega}_{ie}^e &= \omega_{ie}^e - \varepsilon_1 g^e - \varepsilon_2 \mathbf{C}_i^e \dot{\mathbf{r}}^i \end{aligned} \quad (26)$$

See Appendix C for further discussions about other feasible expressions of the two trident twists.

Based on Eqs. (25)-(26), it can be seen that the whole strapdown INS kinematics are encompassed in one elegant trident quaternion differential equation, which takes the same form as the traditional attitude kinematic equation in Eq. (49) and the dual quaternion kinematic equation in Eq. (17). Table I summarizes the aforementioned representations for easy reference.

TABLE I
SUMMARY OF KINEMATIC REPRESENTATIONS OF STRAPDOWN INS

| Methods | Representations |
|---|---|
| Traditional representation | $\dot{\mathbf{C}}_b^e = \mathbf{C}_b^e (\boldsymbol{\omega}_{ib}^b \times) - (\boldsymbol{\omega}_{ie}^e \times) \mathbf{C}_b^e$ <br> $\dot{\mathbf{v}}^e = \mathbf{C}_b^e \mathbf{f}^b - 2(\boldsymbol{\omega}_{ie}^e \times) \mathbf{v}^e + \mathbf{g}_l^e$ <br> $\dot{\mathbf{r}}^e = \mathbf{v}^e$ |
| Dual quaternion representation (DQ) | $2\dot{\hat{q}}_{IT} = \hat{q}_{IT} \circ \hat{\omega}_{IT}^T$ <br> $2\dot{\hat{q}}_{IG} = \hat{q}_{IG} \circ \hat{\omega}_{IG}^G$ <br> $2\dot{\hat{q}}_{IU} = \hat{q}_{IU} \circ \hat{\omega}_{IU}^U$ |
| Dual quaternion and vector hybrid representation (DQv) | $2\dot{\hat{q}}_{ev} = \hat{q}_{ev} \circ \hat{\omega}_{ib}^b - \hat{\omega}_{ie}^e \circ \hat{q}_{ev}$ <br> $\dot{\mathbf{r}}^e = 2q_{ev}' \circ q_{ev}^* - (\boldsymbol{\omega}_{ie}^e \times) \mathbf{r}^e$ |
| Trident quaternion representation (TriQ) | $2\dot{\breve{q}}_{eb} = \breve{q}_{eb} \circ \breve{\omega}_{ib}^b - \breve{\omega}_{ie}^e \circ \breve{q}_{eb}$ |

In Eq. (23), we actually applied the trident quaternion to represent the spatial motion of a rigid body, including the attitude, velocity and position relative to a reference frame, which is also termed as the extended pose on $SE_2(3)$ in [41]. The intrinsic reason for this correspondence is that the set of trident quaternions is innately a Lie group as proved in the following theorem.

*Theorem 1:* A set of unit trident quaternions, denoted as $TriQ$, is a Lie group under the trident quaternion multiplication.

*Proof:* According to the properties of a group, we have:
(1) The closure is obvious in that for any $\breve{q}_1, \breve{q}_2 \in TriQ$, we have the third equation in Eq. (21) and $q_1 \circ q_2$, $q_1 \circ q_2' + q_1' \circ q_2$ and $q_1 \circ q_2'' + q_1'' \circ q_2$ are usual quaternions. Therefore, the closure is ensured by the multiplication $\breve{q}_1 \circ \breve{q}_2 \in TriQ$.

(2) The associate of any $\breve{q}_1, \breve{q}_2, \breve{q}_3 \in TriQ$ can be easily verified as $(\breve{q}_1 \circ \breve{q}_2) \circ \breve{q}_3 = \breve{q}_1 \circ (\breve{q}_2 \circ \breve{q}_3)$.

(3) The identity element is $1 = [1, \mathbf{0}] + \varepsilon_1 [0, \mathbf{0}] + \varepsilon_2 [0, \mathbf{0}]$ by definition.

(4) The inverse is of a unit trident quaternion is $\breve{q}^{-1} = \breve{q}^*$.

Properties (1)-(4) indicate that the set of trident quaternions is a group. It is also known that any unit quaternion $q$ is a differential manifold of three dimensions [26]. Thus, any $\breve{q} \in TriQ$ is a manifold of nine dimensions since its real and imaginary parts are usual quaternions. Besides, for any $\breve{q}_1, \breve{q}_2 \in TriQ$, $\mu(\breve{q}_1, \breve{q}_2) = \breve{q}_1^{-1} \circ \breve{q}_2$ is a smooth (differentiable) mapping in terms of multiplications. As a result, a set of trident quaternions form a Lie group under the trident quaternion multiplication. Proof ends here.

Note that an alternative representation of the trident quaternion can be given as

$$\breve{q}_{eb} = q_{eb} + \varepsilon_1 \frac{1}{2} q_{eb} \circ (\mathbf{C}_i^b \dot{\mathbf{r}}^i) + \varepsilon_2 \frac{1}{2} q_{eb} \circ (\mathbf{C}_i^b \mathbf{r}^i) \quad (27)$$

which can be obtained from Eq. (23) by the following relationship

$$(\mathbf{C}_i^e \dot{\mathbf{r}}^i) \circ q_{eb} = q_{eb} \circ q_{eb}^* \circ (\mathbf{C}_i^e \dot{\mathbf{r}}^i) \circ q_{eb} = q_{eb} \circ (\mathbf{C}_i^b \dot{\mathbf{r}}^i) \quad (28)$$

Moreover, the trident quaternion can be reversely defined as the $e$-frame relative to the $b$-frame, which is actually the robocentric formulations as in [41], [42]. We have

$$\breve{q}_{be} = q_{be} + \varepsilon_1 \frac{1}{2} (\mathbf{C}_i^b \dot{\mathbf{r}}^i) \circ q_{be} + \varepsilon_2 \frac{1}{2} (\mathbf{C}_i^b \mathbf{r}^i) \circ q_{be} \quad (29)$$

Or

$$\breve{q}_{be} = q_{be} + \varepsilon_1 \frac{1}{2} q_{be} \circ (\mathbf{C}_i^e \dot{\mathbf{r}}^i) + \varepsilon_2 \frac{1}{2} q_{be} \circ (\mathbf{C}_i^e \mathbf{r}^i) \quad (30)$$

The differential equations of Eqs. (29) and (30) can be easily computed as in Eq. (24). Of course, the trident quaternions (29) and (30) are just conjugate counterparts of (23) and (27), with opposite signs in the imaginary parts.

## V. Functional Iterative Integration of Trident Quaternion Kinematics

This section devotes to solve the trident quaternion differential equation by the functional iterative integration framework proposed in [44], named as the tqFIter algorithm hereafter. The trident quaternion kinematic equation based on the first specific solution in Eq. (63) is repeatedly given here, omitting subscripts of the trident quaternion for symbolic brevity in the subsequent development.

$$2\dot{\breve{q}} = \breve{q} \circ \breve{\omega}_{ib}^b - \breve{\omega}_{ie}^e \circ \breve{q} \quad (31)$$

where $\breve{q} = q + \varepsilon_1 q' + \varepsilon_2 q''$. Notice that the velocity and position can be readily recovered from Eqs. (23) and (11)

$$\begin{aligned} \mathbf{r}^e &= 2q'' \circ q^* \\ \mathbf{v}^e &= 2q' \circ q^* - \boldsymbol{\omega}_{ie}^e \times \mathbf{r}^e \end{aligned} \quad (32)$$

According to the functional iterative iteration approach, integrating the differential equation in Eq. (31) over the time interval $[0, t]$ gives

$$\breve{q} = \breve{q}(0) + \int_0^t (\breve{q} \circ \breve{\omega}_{ib}^b - \breve{\omega}_{ie}^e \circ \breve{q})/2 \, dt \quad (33)$$

According to Eqs. (23) and (32), the initial value of the trident quaternion is given as

$$\breve{q}(0) = \breve{q}|_{t=0} = \left( q + \varepsilon_1 \frac{1}{2} (\mathbf{v}^e + \boldsymbol{\omega}_{ie}^e \times \mathbf{r}^e) \circ q + \varepsilon_2 \frac{1}{2} \mathbf{r}^e \circ q \right)\bigg|_{t=0} \quad (34)$$

Without loss of generality, we will consider the navigation updates over the time interval $[0, t]$, in which $N$ samples of triads of gyroscopes and accelerometers are available. At time instants $t_k$ ($k = 1, 2, \ldots N$), assume the discrete angular velocity $\tilde{\boldsymbol{\omega}}_{t_k}$ or the angular increment (integrated angular velocity) $\Delta\tilde{\boldsymbol{\theta}}_{t_k}$ measurements by a triad of gyroscopes, and the discrete specific force $\tilde{\mathbf{f}}_{t_k}$ or velocity increments (integrated specific force) $\Delta\tilde{\mathbf{v}}_{t_k}$ measurements by a triad of accelerometers are available. In order to apply the Chebyshev polynomials, the actual time interval is mapped onto $[-1, 1]$ by letting $t = t_N(1 + \tau)/2$.

Assume the trident quaternion estimate at the $l$-th iteration is expressed by a weighted sum of Chebyshev polynomials, say

$$\breve{q}_l = \sum_{i=0}^{m_q} \breve{b}_{l,i} F_i(\tau) \quad (35)$$

where $m_q$ denotes the maximum degree of Chebyshev polynomials and $\breve{b}_{l,i}$ is the coefficient of $i^{\text{th}}$-degree Chebyshev polynomial at the $l$-th iteration. $F_i(x)$ is the $i^{\text{th}}$-degree Chebyshev polynomial of the first kind. For any $j, k \geq 0$, it satisfies the equality [49]

$$F_j(\tau) F_k(\tau) = \frac{1}{2} \left( F_{j+k}(\tau) + F_{|j-k|}(\tau) \right) \quad (36)$$

According to the integral property of the Chebyshev polynomial, we have

$$G_{i,[\tau_{k-1}\tau_k]} = \int_{\tau_{k-1}}^{\tau_k} F_i(\tau) d\tau$$

$$= \begin{cases} \left( \frac{iF_{i+1}(\tau_k)}{i^2-1} - \frac{\tau_k F_i(\tau_k)}{i-1} \right) - \left( \frac{iF_{i+1}(\tau_{k-1})}{i^2-1} - \frac{\tau_{k-1} F_i(\tau_{k-1})}{i-1} \right), & i \neq 1 \\ \dfrac{\tau_k^2 - \tau_{k-1}^2}{2}, & i = 1 \end{cases}$$

(37)

In specific, the integrated $i^{\text{th}}$-degree Chebyshev polynomial can be expressed as a linear combination of $(i+1)^{\text{th}}$-degree Chebyshev polynomials, given by

$$G_{i,[-1\ \tau]} = \int_{-1}^{\tau} F_i(\tau) d\tau$$

$$= \begin{cases} \left(\dfrac{iF_{i+1}(\tau)}{i^2-1} - \dfrac{\tau F_i(\tau)}{i-1}\right) - \left(\dfrac{iF_{i+1}(-1)}{i^2-1} + \dfrac{F_i(-1)}{i-1}\right) F_0(\tau) \\ = \left(\dfrac{iF_{i+1}(\tau)}{i^2-1} - \dfrac{F_{i+1}(\tau)+F_{|i-1|}(\tau)}{2(i-1)}\right) - \left(\dfrac{iF_{i+1}(-1)}{i^2-1} + \dfrac{F_i(-1)}{i-1}\right) F_0(\tau) \\ = \left(\dfrac{F_{i+1}(\tau)}{2(i+1)} - \dfrac{F_{|i-1|}(\tau)}{2(i-1)}\right) - \dfrac{(-1)^i}{i^2-1} F_0(\tau) \quad \text{for } i \neq 1, \\ \dfrac{\tau^2-1}{2} = \dfrac{F_{i+1}(\tau)}{4} - \dfrac{F_0(\tau)}{4} \quad \text{for } i = 1, \end{cases}$$

(38)

The fitted trident twists (angular velocity and specific force) using the Chebyshev polynomials can be respectively written as [44]

$$\breve{\omega}_{ib}^b = \sum_{i=0}^{n_{\omega b}} \breve{c}_i F_i(\tau), \quad n_{\omega b} \leq N-1 \tag{39}$$

and,

$$\breve{\omega}_{ie,l}^e = \sum_{i=0}^{n_{\omega e}} \breve{d}_{l,i} F_i(\tau), \quad n_{\omega e} \leq N-1 \tag{40}$$

where $n_{\omega b}$ and $n_{\omega e}$ denote the maximum degrees of Chebyshev polynomials. The coefficient $\breve{c}_i$ are determined by solving the least-square equations of the discrete gyroscope/accelerometer measurements and the coefficient. As the terms $\mathbf{g}^e$ nonlinearly depends on the trident quaternion, $\breve{d}_{l,i} \triangleq d_l + \varepsilon_1 d'_{l,i} + \varepsilon_2 d''_l$ should be approximated by a process given by [49], namely,

$$\begin{aligned} d_l &= \omega_{ie}^e \\ d'_{l,i} &\approx -\frac{2-\delta_{0i}}{P} \sum_{k=0}^{P-1} \cos\frac{i(k+1/2)\pi}{P} \mathbf{g}^e(\breve{q}_l) \\ d''_l &= -2q'_l \circ q_l^* \end{aligned} \tag{41}$$

Using Eq. (35), the third equality in the above equation can be expressed as

$$\begin{aligned} d''_l &= -2\sum_{j=0}^{m_q} b'_{l,j} F_j(\tau) \circ \sum_{k=0}^{m_q} b^*_{l,k} F_k(\tau) \\ &= -\sum_{j=0}^{m_q} \sum_{k=0}^{m_q} b'_{l,j} \circ b^*_{l,k} \left(F_{j+k}(\tau) + F_{|j-k|}(\tau)\right) \end{aligned} \tag{42}$$

The integral in Eq. (33) is transformed to that over the mapped interval of Chebyshev polynomials, that is,

$$\breve{q}_{l+1} = \breve{q}(0) + \frac{t_N}{4} \int_{-1}^{\tau} \left(\breve{q}_l \circ \breve{\omega}_{ib}^b - \breve{\omega}_{ie,l}^e \circ \breve{q}_l\right) d\tau \tag{43}$$

Substitute Eq. (35)-(40) to (43), the numerical solution of the trident quaternion is obtained as

$$\begin{aligned} \breve{q}_{l+1} &= \breve{q}(0) + \frac{t_N}{4} \int_{-1}^{\tau} \left(\breve{q}_l \circ \breve{\omega}_{ib}^b - \breve{\omega}_{ie,l}^e \circ \breve{q}_l\right) d\tau \\ &= \breve{q}(0) + \frac{t_N}{4} \int_{-1}^{\tau} \left(\begin{bmatrix}\sum_{i=0}^{m_q} \breve{b}_{l,i} F_i(\tau)\end{bmatrix} \circ \sum_{j=0}^{n_{\omega e}} \breve{c}_j F_j(\tau) \\ - \begin{bmatrix}\sum_{j=0}^{n_{\omega e}} \breve{d}_{l,j} F_j(\tau)\end{bmatrix} \circ \begin{bmatrix}\sum_{i=0}^{m_q} \breve{b}_{l,i} F_i(\tau)\end{bmatrix}\right) d\tau \\ &= \breve{q}(0) + \frac{t_N}{4} \begin{pmatrix}\sum_{i=0}^{m_q} \sum_{j=0}^{n_{\omega e}} \breve{b}_{l,i} \circ \breve{c}_j \int_{-1}^{\tau} F_i(\tau) F_j(\tau) d\tau \\ - \sum_{j=0}^{n_{\omega e}} \sum_{i=0}^{m_q} \breve{d}_{l,j} \circ \breve{b}_{l,i} \int_{-1}^{\tau} F_j(\tau) F_i(\tau) d\tau\end{pmatrix} \\ &= \breve{q}(0) + \frac{t_N}{8} \begin{pmatrix}\sum_{i=0}^{m_q} \sum_{j=0}^{n_{\omega e}} \breve{b}_{l,i} \circ \breve{c}_j \left(G_{i+j,[-1\ \tau]} + G_{|i-j|,[-1\ \tau]}\right) \\ - \sum_{j=0}^{n_{\omega e}} \sum_{i=0}^{m_q} \breve{d}_{l,j} \circ \breve{b}_{l,i} \left(G_{i+j,[-1\ \tau]} + G_{|i-j|,[-1\ \tau]}\right)\end{pmatrix} \\ &= \sum_{i=0}^{\max(n_{\omega b}, n_{\omega e})+m_q+1} \breve{b}_{l+1,i} F_i(\tau) \overset{\text{polynomial truncation}}{\approx} \sum_{i=0}^{m_q} \breve{b}_{l+1,i} F_i(\tau) \end{aligned}$$

(44)

In contrast with [44], where the attitude computation precedes the velocity/position computation, the current computation involves only one iterative computation process, thanks to the compact trident quaternion representation.

## VI. NUMERICAL RESULTS

Numerical tests are conducted to verify the representation development and the tqFIter algorithm. Note that the simulation data were generated analytically in the local-level frame and then transformed to the earth frame [48]. The vehicle flies around the surface at the ground velocity $\dot{\mathbf{v}}^n = [0, 0, a\sin(wt)]^T$ with an initial east velocity $v_0 = 500\text{m/s}$, in which the magnitude of acceleration $a = 10\text{m/s}^2$, and the angular frequency of the varying acceleration $w = 0.02\pi\text{rad/s}$. The vehicle is initially located at zero longitude, zero latitude and zero height. In addition, the body frame performs a classical coning motion with the attitude quaternion

$$q_{nb} = \cos(\alpha/2) + \sin(\alpha/2)[0\ \cos(\zeta t)\ \sin(\zeta t)]^T \tag{45}$$

which denotes the attitude of the body frame w.r.t. the local-level navigation frame. The coning angle $\alpha = 10\deg$ and the coning frequency $\zeta = 0.74\pi\text{rad/s}$.

The true velocity and position are obtained as

$$\begin{aligned} \mathbf{v}^n &= \mathbf{v}^n(0) + \int_0^t \dot{\mathbf{v}}^n dt = \begin{bmatrix}0 & 0 & v_0 - (a\cos(wt) - a)/w\end{bmatrix}^T \\ \mathbf{p}^n &= \int_0^t \mathbf{R}_c \mathbf{v}^n dt = \begin{bmatrix}v_0 t - (a\sin(wt) - awt)/w^2 & 0 & 0\end{bmatrix}^T \bigg/ R_E \end{aligned}$$

(46)

where $\mathbf{R}_c$ is a function of the current position, the transverse radius of curvature $R_E$ and the meridian radius of curvature $R_N$.

$$\mathbf{R}_c = \begin{bmatrix} 0 & 0 & 1/\left[(R_E+h)\cos L\right] \\ 1/(R_N+h) & 0 & 0 \\ 0 & 1 & 0 \end{bmatrix} \quad (47)$$

Finally, the gyroscope and accelerometer outputs at 100Hz sampling rate are generated according to

$$\begin{aligned} \omega_{ib}^b &= q_{nb}^* \circ \left(2\dot{q}_{nb} + \omega_{in}^n \circ q_{nb}\right) \\ \mathbf{f}^b &= \mathbf{C}_n^b \left(\dot{\mathbf{v}}^n + \left(2\omega_{ie}^n + \omega_{en}^n\right) \times \mathbf{v}^n - \mathbf{g}^n\right) \end{aligned} \quad (48)$$

In this numerical test, $N = 8$ samples of gyroscopes and accelerometers are used to fit the angular velocity and specific force. The orders of Chebyshev polynomials involved in the fitting are set as $n_{\omega b} = n_{\omega e} = N-1$. Besides, the orders of truncation in approximating the attitude quaternion, the first imaginary part and the second imaginary part are all set to $m_q = N+1$. The maximum iteration number in solving Eq. (44) is set as $N+1$. Then, the root-mean-square discrepancy of polynomial coefficients between two iterations is adopted as the termination criterion, which is set to $10^{-16}$. Figures 3-5 show the computation errors of the attitude angle, velocity, and position in 200 seconds. Comparing with the routine two-sample algorithm [5], [6], the tqFIter algorithm (dotted lines) significantly outperforms the traditional method (dashed lines) under the considered scenario. In specific, the attitude/velocity/position computation accuracy is about 8 orders higher than the two-sample algorithm. Note that this accuracy is comparable to the iNavFIter algorithm based on the traditional representation of strapdown INS, albeit the iNavFIter is somewhat more accurate in position computation as shown in Fig. 5. These results show the feasibility and effectiveness of applying the functional iterative integration framework to numerically solve the newly developed trident quaternion kinematics. It should be useful for not only pure inertial navigation but kinematics forward propagation in inertial-based navigation with aiding sensors.

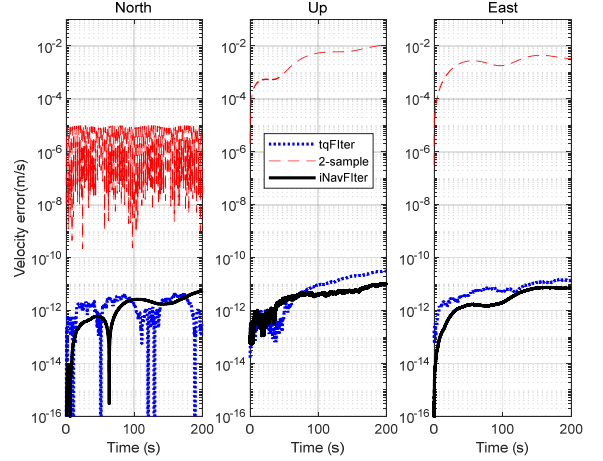

Figure 4: Velocity errors in local-level frame (north, up, east). Dashed line denotes the results of two-sample algorithm, and the dotted lines denotes those of the tqFIter algorithm. Solid lines denote the results of the iNavFIter.

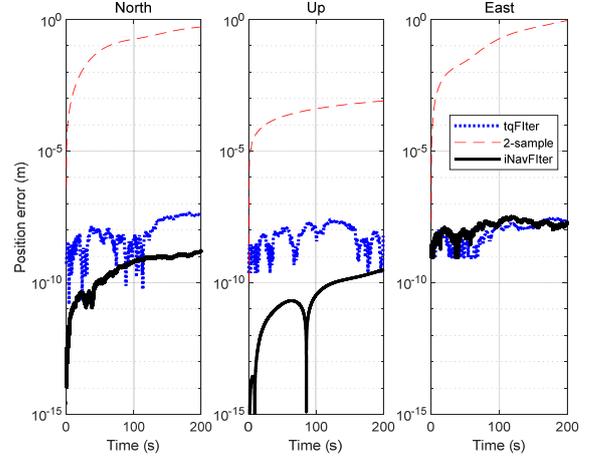

Figure 5: Position errors in local-level frame (north, up, east). Dashed line denotes the results of two-sample algorithm, and the dotted lines denotes those of the tqFIter algorithm. Solid lines denote the results of the iNavFIter.

## VII. CONCLUSIONS

In this article, the kinematic representation of the strapdown INS is investigated. In view of the redundancy existing in the dual quaternion representation, a new total velocity frame is defined to simplify it. The strapdown INS kinematics are subsequently described by a dual quaternion with a position vector. In order to achieve the most concise representation, trident numbers and trident quaternions are devised, which successfully incorporate the position vector into the algebraic representation. Interestingly, the kinematics of the whole strapdown inertial navigation are elegantly described by one unique trident quaternion differential equation. Moreover, we numerically solve the trident quaternion kinematic equation by using the functional iterative integration approach. The proposed trident quaternion model not only provides the most compact representation of strapdown INS, but contributes to the state estimation of inertial-based navigation with aiding sensors, which is further covered in the companion paper.

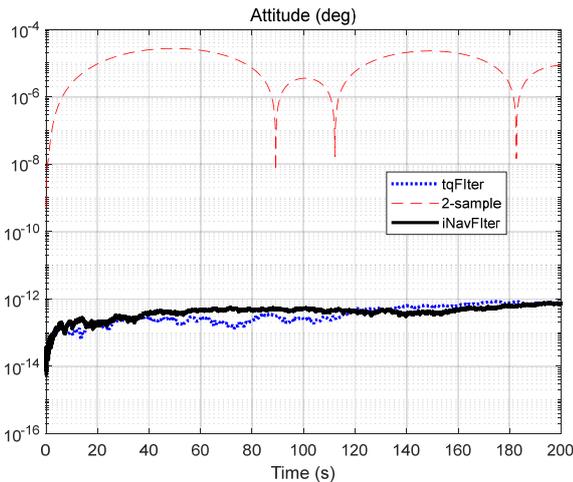

Figure 3: Principal angle errors. Dashed line denotes the results of two-sample algorithm, and the dotted line denotes those of the tqFIter algorithm. The solid line denotes the results of the iNavFIter.

## APPENDIX A

The derivation of Eq. (17) is provided here. The derivative of right-side scalar part in Eq. (16) is

$$2\dot{q}_{eb} = q_{eb} \circ \omega_{eb}^b = q_{eb} \circ \omega_{ib}^b - \omega_{ie}^e \circ q_{eb} \quad (49)$$

Then, the derivative of the imaginary part of Eq. (16) is

$$\begin{aligned}
&\left(\dot{\mathbf{C}}_i^e \dot{\mathbf{r}}^i + \mathbf{C}_i^e \ddot{\mathbf{r}}^i\right) \circ q_{eb} + \left(\mathbf{C}_i^e \dot{\mathbf{r}}^i\right) \circ \dot{q}_{eb} \\
&= -\boldsymbol{\omega}_{ie}^e \times \left(\mathbf{C}_i^e \dot{\mathbf{r}}^i\right) \circ q_{eb} + \mathbf{C}_i^e \left(\mathbf{f}^i + \mathbf{g}^i\right) \circ q_{eb} \\
&\quad + \frac{1}{2}\left(\mathbf{C}_i^e \dot{\mathbf{r}}^i\right) \circ \left(q_{eb} \circ \omega_{ib}^b - \omega_{ie}^e \circ q_{eb}\right) \\
&= -\frac{1}{2}\left(\omega_{ie}^e \circ \left(\mathbf{C}_i^e \dot{\mathbf{r}}^i\right) - \left(\mathbf{C}_i^e \dot{\mathbf{r}}^i\right) \circ \omega_{ie}^e\right) \circ q_{eb} + \left(\mathbf{C}_b^e \mathbf{f}^b\right) \circ q_{eb} \\
&\quad + g^e \circ q_{eb} + \frac{1}{2}\left(\mathbf{C}_i^e \dot{\mathbf{r}}^i\right) \circ \left(q_{eb} \circ \omega_{ib}^b - \omega_{ie}^e \circ q_{eb}\right) \\
&= \frac{1}{2}\left(\mathbf{C}_i^e \dot{\mathbf{r}}^i\right) \circ q_{eb} \circ \omega_{ib}^b + q_{eb} \circ f^b + g^e \circ q_{eb} - \frac{1}{2}\omega_{ie}^e \circ \left(\mathbf{C}_i^e \dot{\mathbf{r}}^i\right) \circ q_{eb}
\end{aligned} \quad (50)$$

In the second step of Eq. (50), the following relationship between two vector quaternions $v_1 = [0, \mathbf{v}_1]$ and $v_2 = [0, \mathbf{v}_2]$ is used.

$$v_1 \times v_2 = \frac{1}{2}(v_1 \circ v_2 - v_2 \circ v_1) = [0, \mathbf{v}_1 \times \mathbf{v}_2] \quad (51)$$

Based on Eq. (49), (50), the result is obtained in Eq. (17).

## APPENDIX B

This appendix provides another perspective to derive Eq. (16) from [23] yet by a minor adjustment to the original frame definitions in contrast with the current work. Note that the symbols may be slightly different from the original ones.

In [23], the *thrust velocity* is defined as the sum of the initial velocity of the vehicle and integration of the specific force $\mathbf{s}^i$, which is denoted in the inertial frame.

$$\mathbf{v}_t^i = \dot{\mathbf{r}}^i(0) + \int_0^t \mathbf{s}^i d\tau \quad (52)$$

Further define the *gravitational velocity* as minus integration of the gravitational acceleration,

$$\mathbf{v}_g^i = -\int_0^t \mathbf{g}^i d\tau \quad (53)$$

Then, the velocity of the IMU in *i*-frame is expressed as

$$\dot{\mathbf{r}}^i = \dot{\mathbf{r}}^i(0) + \int_0^t \left(\mathbf{s}^i + \mathbf{g}^i\right) dt = \mathbf{v}_t^i - \mathbf{v}_g^i \quad (54)$$

Firstly, the thrust velocity frame (*t*-frame) is defined to align with *b*-frame in orientation but displaced relative to *i*-frame by the thrust velocity $\mathbf{v}_t^i$, and therefore $q_{it} = q_{ib}$. Suppose a unit dual quaternion $\hat{q}_{it} = q_{it} + \varepsilon q_{it}' = q_{it} + \varepsilon \frac{1}{2} v_t^i \circ q_{it}$ characterizes the general displacement of *t*-frame relative to *i*-frame, where $v_t^i = [0, \mathbf{v}_t^i]$. According to [23], the twist expressed in *t*-frame is given by

$$\hat{\omega}_{it}^t = \omega_{ib}^b + \varepsilon f^b \quad (55)$$

where the specific force actually satisfies $\mathbf{f}^b = q_{ib}^* \circ \mathbf{s}^i \circ q_{ib}$ and $f^b = [0, \mathbf{f}^b]$. The kinematic equation of $\hat{q}_{it}$ is given as

$$2\dot{\hat{q}}_{it} = \hat{q}_{it} \circ \hat{\omega}_{it}^t \quad (56)$$

Secondly, the *gravitational velocity* frame (*g*-frame) is defined to align with *e*-frame in orientation but displaces relative to *i*-frame by the gravitational velocity. Let the general displacement of *g*-frame relative to *e*-frame be represented by the other unit dual quaternion $\hat{q}_{ig} = q_{ig} + \varepsilon q_{ig}' = q_{ig} + \varepsilon \frac{1}{2} v_g^i \circ q_{ig}$, where $v_g^i = [0, \mathbf{v}_g^i]$. The twist expressed in *g*-frame is

$$\hat{\omega}_{ig}^g = \omega_{ie}^e - \varepsilon g^e \quad (57)$$

The kinematic equation of $\hat{q}_{ig}$ is

$$2\dot{\hat{q}}_{ig} = \hat{q}_{ig} \circ \hat{\omega}_{ig}^g \quad (58)$$

Let $\hat{q}_{gt} = \hat{q}_{ig}^* \circ \hat{q}_{it}$ characterize the relative motion of *t*-frame with respect to *g*-frame. With Eq. (54), then it can be shown that

$$\begin{aligned}
\hat{q}_{gt} &= q_{ig}^* \circ q_{it} + \varepsilon \left(q_{ig}'^* \circ q_{it} + q_{ig}^* \circ q_{it}'\right) \\
&= q_{gt} + \frac{1}{2}\varepsilon \left(q_{ig}^* \circ v_t^i \circ q_{it} - q_{ig}^* \circ v_g^i \circ q_{it}\right) \\
&= q_{gt} + \frac{1}{2}\varepsilon q_{ig}^* \circ \left(v_t^i - v_g^i\right) \circ q_{it} \\
&= q_{gt} + \frac{1}{2}\varepsilon \left(\mathbf{C}_i^e \dot{\mathbf{r}}^i\right) \circ q_{gt}
\end{aligned} \quad (59)$$

which is exactly equal to Eq. (16) because *g*-frame and *t*-frame share the same orientations with *e*-frame and *b*-frame, respectively. That is to say, $\hat{q}_{gt} \equiv \hat{q}_{eb}$. Finally, the kinematics of $\hat{q}_{gt}$ satisfy

$$\begin{aligned}
2\dot{\hat{q}}_{gt} &= 2\hat{q}_{ig}^* \circ \dot{\hat{q}}_{it} - 2\dot{\hat{q}}_{ig}^* \circ \hat{q}_{ig} \circ \hat{q}_{ig}^* \circ \hat{q}_{it} \\
&= \hat{q}_{ig}^* \circ \hat{q}_{it} \circ \hat{\omega}_{it}^t - \hat{q}_{ig}^* \circ \hat{q}_{ig} \circ \hat{\omega}_{ig}^g \circ \hat{q}_{ig}^* \circ \hat{q}_{it} \\
&= \hat{q}_{gt} \circ \hat{\omega}_{it}^t - \hat{\omega}_{ig}^g \circ \hat{q}_{gt}
\end{aligned} \quad (60)$$

which is just the Eq. (17) in the current work.

## APPENDIX C

Actually, the expressions of two trident twists can be assumed as the following general format

$$\begin{aligned}
\breve{\omega}_{ib}^b &= \omega_{ib}^b + \varepsilon_1 f^b + \varepsilon_2 x_1 \\
\breve{\omega}_{ie}^e &= \omega_{ie}^e - \varepsilon_1 g^e + \varepsilon_2 x_2
\end{aligned} \quad (61)$$

where $x_1, x_2$ are vector quaternions.

Substitute Eq. (61) into Eq. (25) and compare with Eq. (24), the following condition should be satisfied

$$\left(q_{eb} \circ x_1 - x_2 \circ q_{eb}\right) = \left(\mathbf{C}_i^e \dot{\mathbf{r}}^i\right) \circ q_{eb} \quad (62)$$

Obviously, two special solutions of this equation are

$$x_1 = [0, \mathbf{0}], \ x_2 = [0, -\mathbf{C}_i^e \dot{\mathbf{r}}^i] \quad (63)$$

and,

$$x_1 = [0, \mathbf{C}_i^b \dot{\mathbf{r}}^i], \ x_2 = [0, \mathbf{0}] \quad (64)$$

where $\mathbf{0}$ is the three-dimensional vector.

This work adopts the first solution to represent the kinematic model and more discussions about two specific solutions are to be made in the companion work.

## References

[1] D. A. Tazartes, Inertial navigation: From gimbaled platforms to strapdown sensors. *IEEE Trans. Aerosp. Electron. Syst.*, vol. 47, no. 3, pp. 2292–2299, 2010.

[2] J. W. Jordan, An accurate strapdown direction cosine algorithm NASA TN-D-5384, 1969.

[3] J. E. Bortz, A new mathematical formulation for strapdown inertial navigation. *IEEE Trans. Aerosp. Electron. Syst.*, vol. 7, no. 1, pp. 61–66, Jan. 1971.

[4] M. B. Ignagni, Optimal strapdown attitude integration algorithms. *J. Guid., Control, Dyn.*, vol. 13, pp. 363–369, 1990.

[5] P. G. Savage, Strapdown inertial navigation integration algorithm design, Part 1: Attitude algorithms. *J. Guid., Control, Dyn.*, vol. 21, pp. 19–28, 1998.

[6] P. G. Savage, Strapdown inertial navigation integration algorithm design, Part 2: Velocity and position algorithms. *J. Guid., Control, Dyn.*, vol. 21, pp. 208–221, 1998.

[7] P. G. Savage, A unified mathematical framework for strapdown algorithm design. *J. Guid. Control Dyn.*, vol. 29, pp. 237–249, 2006.

[8] D. H. Titterton, J. L. Weston, *Strapdown Inertial Navigation Technology*, 2nd ed.: the Institute of Electrical Engineers, London, United Kingdom, 2007.

[9] P. D. Groves, *Principles of GNSS, Inertial, and Multisensor Integrated Navigation Systems*, 2nd ed.: Artech House, Boston and London, 2013.

[10] M. D. Shuster, A survey of attitude representations. *Journal of the Astronautical Sciences*, vol. 41, no. 4, pp. 439-517, 1993.

[11] F. L. Markley, J. L. Crassidis, *Fundamentals of Spacecraft Attitude Determination and Control*. New York, USA, Springer, 2014.

[12] Y. Wu, RodFIter: Attitude Reconstruction from Inertial Measurement by Functional Iteration, *IEEE Trans. on Aerospace and Electronic Systems*, vol. 54, pp. 2131-2142, 2018.

[13] Y. Wu, Q. Cai, T.-K. Truong, Fast RodFIter for Attitude Reconstruction from Inertial Measurement, *IEEE Trans. on Aerospace and Electronic Systems*, vol. 55, pp. 419-428, 2019.

[14] E J. Leffens, F. L. Markley, M. D. Shuster, Kalman filtering for spacecraft attitude estimation, *J. Guid. Control Dyn.*, vol. 5, no. 5, pp. 417-429, 1982.

[15] G. Yan, J. Weng, X. Yang, and Y. Qin, An accurate numerical solution for strapdown attitude algorithm based on Picard iteration J. Astronaut., vol. 38, pp. 65–71, 2017.

[16] J. Funda, R. H. Taylor, R. P. Paul, On homogeneous transforms, quaternions, and computational efficiency. *IEEE Transactions on Robotics and Automation*, vol. 6, no. 3, pp. 382-388, 1990.

[17] Y. Wu and G. Yan, Attitude Reconstruction From Inertial Measurements: QuatFIter and Its Comparison with RodFIter, *IEEE Transactions on Aerospace and Electronic Systems*, vol. 55, no. 6, pp. 3629-3639, 2019.

[18] W. R. Hamilton, *Elements of Quaternions*. Cambridge: Cambridge University Press, 2010.

[19] R. S. Ball, A Treatise on the Theory of Screws. New York: Cambridge University Press, 1900.

[20] N. A., Aspragathos, and J. K. Dimitros, A comparative study of three methods for robot kinematics. *IEEE Transactions on Systems, Man and Cybernetics Part B: Cybernetics*, vol. 28, no. 2, pp. 135-145, Apr. 1998.

[21] F. Thomas, Approaching dual quaternions from matrix algebra. *IEEE Transactions on Robotics*, vol. 30, no. 5, 1037-1048, 2014.

[22] V. N. Branets, and I. P. Shmyglevsky, Introduction to the Theory of Strapdown Inertial Navigation System (In Russian). Moscow: Nauka, ch. 1-6, 1992.

[23] Y. Wu, X. Hu, D. Hu, T. Li, J. Lian, Strapdown inertial navigation system algorithms based on dual quaternions, IEEE Transactions on Aerospace and Electronic Systems, vol. 41, pp. 110-132, 2005.

[24] Y. Wu, M. Wu, D. Hu, X. Hu, Strapdown inertial navigation system using dual quaternions: error analyis, IEEE Transactions on Aerospace and Electronic Systems, vol. 42, pp. 259-266, 2006.

[25] P. Tsiotras, A. Valverde, Dual Quaternions as a Tool for Modeling, Control, and Estimation for Spacecraft Robotic Servicing Missions. *The Journal of the Astronautical Sciences*, vol. 67, no. 2, pp. 595-629, 2020.

[26] X. Wang, D. Han, C. Yu, Z. Zheng, The geometric structure of unit dual quaternion with application in kinematic control, *Journal of Mathematical Analysis and Applications*, vol. 389, no. 2, pp. 1352-1364, 2012.

[27] X. Wang, C. Yu, Z. Lin, A dual quaternion solution to attitude and position control for rigid-body coordination. *IEEE Transactions on Robotics*, vol. 28, no. 5, pp. 1162-1170, 2012.

[28] D. Han, Q. Wei, Z. Li, W. Sun, Control of oriented mechanical systems: A method based on dual quaternion. *IFAC Proceedings*, vol. 41, no. 2, pp. 3836-3841, 2008.

[29] N. Filipe, A. Valverde, P. Tsiotras. Pose tracking without linearand angular-velocity feedback using dual quaternions. *IEEE Transactions on Aerospace and Electronic Systems*, vol. 52, no. 1, pp. 411-422, 2016.

[30] N. Filipe, M. Kontitsis, P. Tsiotras. Extended Kalman filter for spacecraft pose estimation using dual quaternions. *Journal of Guidance, Control, and Dynamics*, vol. 38, no. 9, pp. 1625-1641, 2015.

[31] K. Li, F. Pfaff, U. D. Hanebeck, Unscented Dual Quaternion Particle Filter for SE (3) Estimation. *IEEE Control Systems Letters*, vol. 5, no. 2, pp. 647-652, 2020.

[32] R. A. Srivatsan, G. T. Rosen, D. F. N. Mohamed, H. Choset. Estimating SE (3) elements using a dual quaternion based linear Kalman filter. *Robotics: Science and systems*, 2016.

[33] S. Bultmann, K. Li, U. D. Hanebeck, Stereo visual slam based on unscented dual quaternion filtering. *22th IEEE International Conference on Information Fusion (FUSION)* pp. 1-8, July, 2019.

[34] A. Sveier and O. Egeland, Dual Quaternion Particle Filtering for Pose Estimation, *IEEE Transactions on Control Systems Technology*, early access, doi: 10.1109/TCST.2020.3026926.

[35] J. A. Fike, J.J. Alonso, The development of hyper-dual numbers for exact second-derivative calculations. *49th AIAA Aerospace Sciences Meeting including the New Horizons Forum and Aerospace Exposition*. AIAA, July, 2011-886.

[36] J. A. Fike, J.J. Alonso, Automatic differentiation through the use of hyper-dual numbers for second derivatives. *Recent Advances in Algorithmic Differentiation*. Springer, Berlin, Heidelberg, pp. 163-173, 2012.

[37] A. Cohen, M. Shoham, Application of hyper-dual numbers to rigid bodies equations of motion. *Mechanism and Machine Theory*, vol. 111, pp. 76-84, 2017.

[38] A. Cohen, M. Shoham. Hyper Dual Quaternions representation of rigid bodies kinematics. *Mechanism and Machine Theory*, vol. 150, no. 103861, 2020.

[39] F. Messelmi, Multidual numbers and their multidual functions. *Electronic Journal of Mathematical Analysis and Applications*, vol. 3, no. 2, pp. 154-172, 2015.

[40] D. Condurache, Multidual Algebra and Higher-Order Kinematics. *European Conference on Mechanism Science*, Springer, Cham. pp. 48-55, Sept, 2020.

[41] R. Hartley, M. Ghaffari, R. M. Eustice, J.W. Grizzle., Contact-aided invariant extended Kalman filtering for robot state estimation. *The International Journal of Robotics Research*. vol. 39, no. 4, pp. 402-430, 2020.

[42] M. Bloesch, M. Burri, S. Omari, M. Hutter, R. Siegwart. Iterated extended Kalman filter based visual-inertial odometry using direct photometric feedback. *The International Journal of Robotics Research*. Vol. 36, no. 10, pp. 1053-1072, 2017.

[43] G. Huang, Visual-Inertial Navigation: A Concise Review, *2019 International Conference on Robotics and Automation (ICRA)*, Montreal, QC, Canada, 2019, pp. 9572-9582

[44] Y. Wu, iNavFIter: Next-Generation Inertial Navigation Computation Based on Functional Iteration, *IEEE Trans. on Aerospace and Electronic Systems*, vol. 56, pp. 2061-2082, 2020.

[45] M. B. Ignagni, Enhanced Strapdown Attitude Computation, *Journal of Guidance, Control, and Dynamics*, vol. 43, no. 6, pp. 1220-1224, 2020.

[46] M. B. Ignagni, Accuracy Limits on Strapdown Velocity and Position Computations, *Journal of Guidance, Control, and Dynamics*, 1-5, 2020, doi.org/10.2514/1.G005538.

[47] P. G. Savage and , M. B. Ignagni, Honeywell Laser Inertial Navigation System (LINS) Test Results, *IEEE Ninth Joint Services data Exchange Group For Inertial Systems*, St. Petersburg, Florida, 18-19 November 1975.

[48] Y. Wu, Motion Representation and Computation - Inertial Navigation and Beyond. Project in Researchgate: https://www.researchgate.net/project/Motion-Representation-and-Computation-Inertial-Navigation-and-Beyond.

[49] W. H. Press, Numerical Recipes: the Art of Scientific Computing, 3rd ed. Cambridge; New York: Cambridge University Press, 2007.